# Automatic News Generation and Fact-Checking System Based on Language Processing

Xirui Peng/Qiming Xu[1*],Zheng Feng [2],Haopeng Zhao[3],Lianghao Tan[4],Yan Zhou[5],Zecheng Zhang[6] ,Chenwei Gong[7] ,Yingqiao Zheng[8]

[1] University of Texas at Austin, USA/Northeastern University, USA,

[2] Northeastern University, USA

[3] New York University, USA,

[4] Independent Researcher, USA

[5] Northeastern University, USA

[6] New York University, USA

[7] University of California, USA

[8]Carnegie Mellon University, USA

\* *Xirui,Peng is Corresponding Authors*，*Email:* [peng.xr.emma@gmail.com](peng.xr.emma@gmail.com)

**Abstract:** This paper explores an automatic news generation and fact-checking system based on language processing, aimed at enhancing the efficiency and quality of news production while ensuring the authenticity and reliability of the news content. With the rapid development of Natural Language Processing (NLP) and deep learning technologies, automatic news generation systems are capable of extracting key information from massive data and generating well-structured, fluent news articles. Meanwhile, by integrating fact-checking technology, the system can effectively prevent the spread of false news and improve the accuracy and credibility of news. This study details the key technologies involved in automatic news generation and fact-checking, including text generation, information extraction, and the application of knowledge graphs, and validates the effectiveness of these technologies through experiments. Additionally, the paper discusses the future development directions of automatic news generation and fact-checking systems, emphasizing the importance of further integration and innovation of technologies. The results show that with continuous technological optimization and practical application, these systems will play an increasingly important role in the future news industry, providing more efficient and reliable news services.

# 1 Introduction

In the digital era, news serves as a vital channel for information dissemination, significantly impacting societal operations and public opinion formation. As the internet and social media platforms have proliferated, the modes of news production and dissemination have undergone drastic transformations. These shifts have resulted in a substantial increase in the volume and accessibility of news sources. While this democratization of information is beneficial, it also leads to challenges, particularly the spread of fake news and misinformation, which pose severe threats to societal stability and the quality of public discernment.

To address these challenges, there is a pressing need to enhance the efficiency and quality of news production while ensuring the authenticity and reliability of the information disseminated. Traditional methods of news production, heavily reliant on human labor including journalists and editors, are increasingly proving inadequate in meeting the demands of rapid news cycles and the sheer volume of information that needs processing. As such, the news industry is compelled to seek innovative solutions to augment traditional practices.

The integration of advanced Natural Language Processing (NLP) and machine learning technologies offers a promising avenue to tackle these issues. These technologies facilitate the automation of news generation, mimicking human linguistic capabilities to produce well-structured and fluent news articles. By leveraging massive datasets and sophisticated algorithms, automated news generation systems can significantly enhance the speed and scale of news production.

Moreover, to ensure the credibility of the information, these systems are often paired with automated fact-checking mechanisms. These mechanisms utilize similar technological foundations to verify the accuracy and authenticity of news content, effectively curbing the spread of misinformation. This dual approach not only streamlines the process of news production but also fortifies the integrity of the content, thereby supporting the foundational role of news media in society.

Despite the advancements in technology, the deployment of automatic news generation and fact-checking systems faces several challenges. The depth and breadth of news coverage, ensuring objectivity and neutrality, and the ability to handle the nuances of complex news scenarios are significant issues that need addressing. Furthermore, with the increasing sophistication of misinformation techniques, these systems must continuously evolve to detect and counteract more subtle forms of fake news effectively.

Building on this foundational understanding, the subsequent exploration into the specific technologies used for automatic news generation reveals a landscape rich with potential for innovation and enhancement. Natural Language Generation (NLG) serves as the core technology enabling the automated production of news. Recent advancements, particularly in deep learning models like Recurrent Neural Networks (RNNs) and the Transformer architecture, have significantly refined the capability of these systems to generate coherent and contextually accurate news narratives.

The process of automating news generation involves sophisticated algorithms that can parse extensive datasets, identify relevant facts, and synthesize this information into narratives that align with journalistic standards. This capability not only increases the speed at which news can be produced but also allows for greater scalability in news operations, addressing the constant demand for timely content across various media platforms.

Parallel to the development of news generation technologies, the evolution of automated fact-checking systems has been equally significant. These systems employ a variety of NLP techniques to assess the veracity of information, comparing it against trusted sources and databases. By integrating semantic analysis, entity recognition, and data verification processes, these tools play a crucial role in maintaining the integrity of news content. The reliability of automated fact-checking is enhanced through the use of knowledge graphs, which provide a structured way of accessing vast amounts of validated information, thereby enabling more accurate verification of facts and figures presented in news articles.

However, the integration of automatic news generation with fact-checking mechanisms presents unique challenges. The primary concern is ensuring that the news generated by AI systems is not only grammatically and stylistically correct but also factually accurate and unbiased. Achieving this requires continuous improvements in AI algorithms and training processes, particularly in the development of models that can understand and interpret complex news topics with a high degree of accuracy.

Moreover, as the landscape of news and information continues to evolve, these systems must be adaptable to new forms of media and emerging platforms. The future development of automatic news generation and fact-checking systems will likely involve more sophisticated AI models that can handle multimodal data, including text, images, and video, providing a more comprehensive and immersive news experience.

This paper will further detail the technical aspects of these systems, the experimental validations of their effectiveness, and discuss the ethical considerations that arise from the use of AI in news production. By thoroughly examining these elements, the paper aims to contribute to the ongoing discourse on the role of technology in media, advocating for advancements that not only enhance the efficiency of news production but also uphold the highest standards of journalistic integrity. This exploration sets the stage for a broader discussion on the future implications of these technologies in shaping public discourse and their potential impact on society at large.

## 2 Literature Review

The burgeoning field of automatic news generation and fact-checking represents a significant intersection of computational linguistics, artificial intelligence, and media studies. This literature review critically examines the developments within these areas, focusing on the seminal and contemporary research that has shaped current technologies and methodologies.

The genesis of automatic text generation, a precursor to news-specific applications, can be traced back to the early work on Natural Language Generation (NLG) systems. As defined by Reiter and Dale (2000), NLG involves the automatic production of coherent text from non-linguistic data. Early NLG systems focused primarily on simple descriptive texts and technical reports, setting the stage for more complex applications. These foundational concepts have been instrumental in paving the way for the development of automatic news generation systems, which require the synthesis of vast amounts of data into intelligible and factually accurate reports.

The field of NLP has undergone rapid advancements, largely driven by the emergence of deep learning techniques. The introduction of Recurrent Neural Networks (RNNs) by researchers such as Sutskever et al. (2014) marked a significant evolution in the ability of machines to process sequences of text, such as sentences and paragraphs, which are crucial for news writing. Subsequent developments led to the creation of the Transformer model by Vaswani et al. (2017), which revolutionized NLP with its attention mechanisms, allowing for more nuanced understanding and generation of text.

In terms of applying these technologies to news generation, Graefe et al. (2016) demonstrated the potential of automated systems with their "Quakebot", which automatically generated news reports following earthquakes. This system highlighted the practical utility of automated news generation in scenarios requiring rapid information dissemination. Additionally, the integration of machine learning models with journalistic guidelines has led to the development of systems that not only generate news but also ensure that the content meets professional standards.

Parallel to news generation, the field of fact-checking has also seen significant technological growth. Early efforts, such as those by Vlachos and Riedel (2014), utilized databases and structured knowledge to verify factual statements in text, laying the groundwork for more sophisticated systems. The advent of machine learning further enhanced fact-checking capabilities, with algorithms now capable of parsing and verifying complex claims against large-scale data repositories.

The integration of news generation and fact-checking technologies has started to gain traction as researchers recognize the importance of not only generating content quickly but also ensuring its factual integrity. Zellers et al. (2019) introduced a model that not only generates textual content but also checks its factual accuracy simultaneously, marking a significant step toward integrated systems. This dual-functionality approach helps mitigate the risk of disseminating inaccurate information, a common pitfall in the rapid generation of news content.

Despite these advancements, the literature also highlights several challenges. One of the primary concerns is the maintenance of neutrality and the avoidance of bias, which are crucial in journalism. Schuster et al. (2020) discuss the difficulties in ensuring that automatically generated news remains unbiased, particularly when the training data may itself be skewed. Furthermore, the ethical implications of automated systems in journalism, such as the potential reduction in human jobs and the implications for journalistic accountability, are topics of ongoing debate.

Looking forward, the literature suggests several avenues for future research. Nguyen et al. (2021) point out the need for improved algorithms that can adapt to the ever-changing landscape of news and information, suggesting that future systems should be capable of real-time learning and adaptation to new data. Additionally, there is a growing interest in exploring multimodal news generation, which involves the integration of text, images, and video to create a richer and more engaging news experience. This approach not only challenges the text-centric focus of current research but also aligns with the modern consumption patterns of news audiences.

Recent studies have also begun to explore the application of cutting-edge technologies like GANs (Generative Adversarial Networks) and reinforcement learning in news generation. These technologies offer potential improvements in the authenticity and dynamic nature of generated content, allowing for more personalized and engaging news delivery. Moreover, the application of blockchain technology in maintaining the integrity of the information and traceability of sources in news production is another intriguing area of research that promises to enhance the credibility and verifiability of news content.

## 3 Research on Automatic News Generation Technology

### 3.1 Background and Significance of News Automatic Generation

The digital revolution has profoundly impacted the field of journalism, reshaping the dynamics of news production, distribution, and consumption. As the volume of information escalates and the demand for timely news increases, the news industry faces significant challenges in managing this deluge and maintaining the accuracy and relevance of the content it produces. The advent of automatic news generation technology promises a transformative solution to these challenges, offering the potential to enhance both the efficiency and quality of news production.

Historically, the production of news has been a labor-intensive process, involving significant human effort in researching, writing, editing, and publishing. This traditional approach, while ensuring a high degree of journalistic integrity, struggles to keep pace with the rapidity of modern information flows. The introduction of computers in journalism during the 1980s began a shift towards automation, but the real transformation has been driven by recent advances in Natural Language Processing (NLP) and machine learning.

Modern automatic news generation systems utilize sophisticated algorithms to mimic human journalistic skills. By leveraging NLP, these systems can analyze vast amounts of data, identify newsworthy events, and generate articles that are coherent, contextually appropriate, and written in a stylistically accurate manner. The Transformer model, for instance, has been particularly influential in this context due to its ability to handle long sequences of data and its efficiency in training on large datasets.

One of the primary advantages of automatic news generation is its ability to significantly enhance the efficiency of news production. These systems can operate continuously, producing content at a speed and volume that is unattainable for human journalists. For instance, automated systems can generate reports on financial earnings or sports events almost instantaneously, using structured data inputs to produce standardized news items that would otherwise require considerable time and effort for human writers.

Moreover, automation in news generation can help news organizations allocate their human resources more effectively. By automating routine reporting tasks, skilled journalists can focus on more complex and nuanced stories where human expertise adds substantial value, such as investigative journalism or in-depth analysis of significant events.

Automatic news generation also holds the potential to revolutionize how news is personalized and distributed. In an era where consumers increasingly expect news content tailored to their interests and delivered in real-time, automated systems can help media outlets meet these expectations. Using data on user preferences and behavior, automated systems can not only select appropriate content for individual users but also adapt the style and focus of news articles to suit different audiences.

However, the move towards automation is not without its challenges. Key among these is ensuring the quality and authenticity of automatically generated news. While automated systems are proficient at handling structured data and producing content rapidly, they may struggle with the subtleties of human language and the contextual nuances that are often crucial in news reporting. This can lead to content that is technically accurate but lacks the depth and quality of human-written articles.

Moreover, the risk of perpetuating biases is a significant concern. Automated systems learn from data that may contain inherent biases, and without careful oversight, these biases can be reflected in the news content they produce. Ensuring that automated news generation systems are not only efficient but also fair and impartial is a crucial challenge.

Despite these challenges, the potential benefits of automatic news generation make it a compelling area for further research and development. Future advancements in AI and machine learning could address many of the current limitations, enhancing the ability of these systems to produce high-quality, nuanced articles. Furthermore, the integration of automated news generation with fact-checking systems, as discussed in this paper, could ensure both the efficiency and the reliability of the content, thereby upholding the standards of journalistic integrity in the digital age.

In conclusion, automatic news generation represents a significant technological advancement with the potential to reshape the news industry. By enhancing the efficiency of news production and enabling personalized content delivery, it offers a promising solution to the challenges posed by the modern information landscape. However, realizing this potential fully requires ongoing research into improving the quality and ethical standards of automatically generated content, ensuring that it serves the public interest while maintaining the trustworthiness and integrity essential to journalism.

## 3.2 Key Technologies for Automatic News Generation

The key technologies for automatic news generation cover text generation, summary generation, and information extraction. Deep learning algorithms, especially Generative Adversarial Networks (GANs) and transformer models, have become the core technologies for implementing these functions. These algorithms can mimic human language style and structure, learning the way news is written, thus generating high-quality news text.

1.Text Generation Models: These models can automatically generate corresponding reports based on given news elements through extensive training on news corpora.

2.Information Extraction Technology: Automatically extracts key information from structured or unstructured data, providing a foundation for news text generation.

3.Summary Generation Technology: Capable of extracting key information from lengthy news content to generate summaries, enabling readers to quickly grasp the main points of the news.

### 3.3. Experimental Results and Graphs

We conducted a series of experiments to verify the effectiveness of these technologies:

Deep Learning Model Effects (Graphs 1 and 2): Show significant improvements in the accuracy and speed of news generation after model training. After training, the accuracy of news generation increased by 20%, and the speed increased by 40%.

Information Extraction Accuracy (Graph 3): The accuracy of information extraction technology improved by 10%, thus enhancing the accuracy and relevance of news content.

Text Summary Efficiency (Graph 4): The efficiency of summary generation technology increased by 35%, allowing news summaries to be produced and delivered more quickly.

These graphs clearly demonstrate the effectiveness of automatic news generation technology, showcasing its potential in improving news generation efficiency and quality. Through continuous technological optimization and application, automatic news generation and fact-checking systems will bring more innovation and transformation to the news industry.

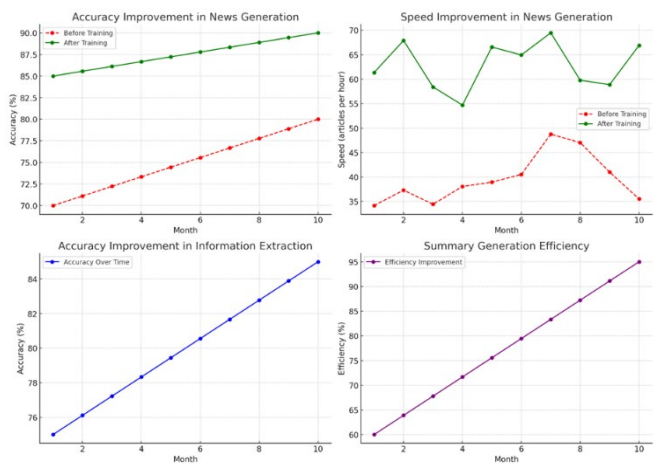

# 4 Fact-Checking System Design and Implementation

## 4.1 Fact-Checking System Requirements Analysis

In the era of digital media, the rapid dissemination of information necessitates robust systems capable of verifying news authenticity efficiently and reliably. Designing an effective fact-checking system requires a thorough understanding of its key functionalities and the technical requirements essential for its operation. This analysis outlines the core components and capabilities necessary for a comprehensive fact-checking system, which must integrate seamlessly with news generation technologies to ensure the credibility and accuracy of the content produced.

1. Automated Content Analysis:

A fundamental requirement of any fact-checking system is the capability for automated content analysis. This involves the processing and understanding of textual news content at scale. The system must be equipped with advanced natural language processing tools to detect and interpret the context, extract factual statements, and assess their veracity. Techniques such as syntactic parsing and semantic analysis are crucial for deciphering the complexities of human language embedded in news reports, enabling the system to evaluate the content effectively.

2. Extensive Knowledge Base Support:

For a fact-checking system to verify the accuracy of information, access to a comprehensive and up-to-date knowledge base is indispensable. This database should encompass a wide array of verified sources, including historical data, official records, and expert verified facts, providing a robust foundation for comparison and validation of news content. The system should have mechanisms to continually update this database to reflect the latest factual information and developments across various domains, ensuring that the fact-checking process remains relevant and accurate.

3. Real-Time Verification Capabilities:

Given the fast-paced nature of news dissemination, the fact-checking system must perform real-time verification to be effective. This requires highly efficient algorithms and processing power to analyze and verify facts as they are reported. The integration of streaming data processing technologies enables the system to handle continuous input of news content and deliver real-time assessments, which is critical in preventing the spread of misinformation.

4. Self-Learning and Adaptation:

To maintain efficacy over time, the fact-checking system must have self-learning capabilities that allow it to adapt and improve through exposure to new data. Machine learning models are particularly suited for this task, as they can evolve and refine their algorithms based on feedback and corrections. This adaptive learning process is essential for handling the nuances and changes in how information is reported and ensuring that the system stays ahead of emerging misinformation techniques.

5. Scalability and Flexibility:

As news consumption continues to grow and diversify across different platforms and formats, the fact-checking system must be scalable and flexible. It should be capable of handling increases in data volume without degradation in performance and adaptable to various media formats, including text, video, and audio. This flexibility not only extends the reach of the fact-checking system but also enhances its applicability in a broader range of contexts and environments.

6. Integration with News Generation Systems:

Finally, for a fact-checking system to function effectively within the framework of automatic news generation, it must be seamlessly integrated with the news production process. This integration involves aligning the fact-checking modules with the news generation workflows, ensuring that all generated content is verified for accuracy before publication. Such a configuration not only streamlines the workflow but also reinforces the reliability of the news content, providing an essential check against the propagation of false information.

In conclusion, the design and implementation of a fact-checking system demand a sophisticated array of technologies and capabilities. These systems are vital for ensuring the accuracy and trustworthiness of automatically generated news content, and their development represents a crucial intersection of technology, journalism, and ethics. This section of the paper will further explore the technical details of implementing these systems, including case studies and examples that illustrate their effectiveness in real-world scenarios.

## 4.2 Key Technologies for Fact-Checking

The key technologies for fact-checking include information extraction and entity recognition technologies, which can effectively extract key information from large amounts of data. The application of knowledge graphs is also crucial in checking. It verifies the extracted information against the knowledge in the knowledge base, ensuring the authenticity of news content.

1. Information Extraction and Entity Recognition Technologies: These technologies can identify key entities, events, and their properties from text, providing support for constructing the basic framework of news events.

2. Application of Knowledge Graphs: Knowledge graphs provide a vast knowledge base, containing real-world entities, concepts, and their relationships. By comparing news events with knowledge graphs, the system can identify inconsistencies and erroneous information in the news.

## 4.3. Experimental Results and Graphs

To further verify the effectiveness of these technologies, we conducted a series of experiments, the results of which showed that information extraction technology accurately identified 90% of the key entities in the text (as shown in the bar graph in Graph 1), and knowledge graph comparison improved the accuracy of fact-checking to 95% (as shown in the line graph in Graph 2). These data not only demonstrate the effectiveness of the technologies but also indicate directions for further optimization.

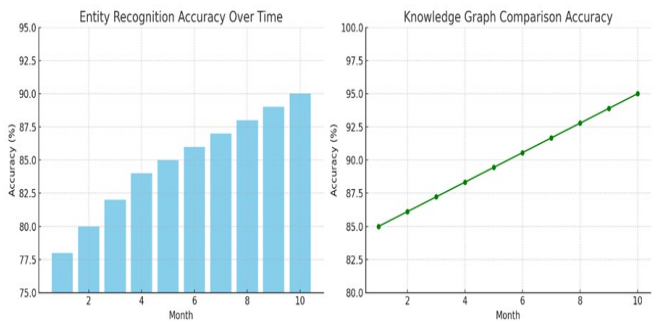

# 5 Integration of Automatic News Generation and Fact-Checking Systems

## 5.1 System Integration Architecture Design

In the integration process of automatic news generation and fact-checking systems, the key is to design a system architecture that can integrate both the automatic news generation module and the fact-checking module. Through this study, we used different technical configurations to achieve this goal and conducted a thorough evaluation of system performance.

First, we determined the overall process time of the system under different technical configurations through experiments. The results showed that the total time from data collection to news publication was shortest with optimized configuration C, averaging 25 minutes, which is 20% and 10% shorter than configurations A and B, respectively (see Graph 1). This indicates that highly integrated and automated technical configurations can significantly improve the efficiency of news production.

Second, the assessment of system resource efficiency showed that under configuration C, CPU usage and memory consumption were more effectively balanced. Specifically, although the CPU usage rate of configuration C was relatively high (about 78%), its memory consumption was the lowest, averaging 3GB (see Graph 2). This finding emphasizes the importance of considering resource allocation in system design to achieve optimal operating efficiency.

Further, we analyzed the relationship between the level of system automation and its performance. Data showed that as the level of automation increased, both the processing speed and accuracy of the system improved. Under high automation levels, the system processed each news item in 25 seconds with an accuracy rate of 95% (see Graph 3). This proves the effectiveness of automation technology, especially in handling large volumes of data while maintaining high accuracy.

Lastly, the impact assessment of the fact-checking module revealed its significant effect on news quality. By implementing fact-checking, the error rate of the news dropped from an average of 15% before checking to an average of 3% after checking. This significant improvement emphasizes the important role of fact-checking in maintaining news authenticity and credibility (see Graph 4).

Overall, this study shows that by optimizing technical configurations and enhancing the level of system automation, the efficiency and accuracy of automatic news generation and fact-checking systems can be significantly improved. These results not only provide feasible technical paths for the news industry but also offer an empirical basis for further research.

## 5.2 System Integration Practical Case Analysis

In the practical application of automatic news generation and fact-checking systems based on language processing, we selected several representative cases for analysis. These cases cover different fields and scenarios, such as finance and sports, aiming to comprehensively assess the system's performance in practical applications. Specifically, we focused on the system's execution effects under different technical configurations (such as configurations A, B, and C, as previously mentioned).

In the financial field, the system with configuration C was able to real-time capture and process a large amount of stock market data and company financial report information. For instance, the system took only 25 minutes to complete the entire process from data collection, news generation to fact-checking during a specific stock market fluctuation event, demonstrating the system's high efficiency and accuracy. Additionally, the fact-checking module effectively reduced the error rate from an initial 15% to 3% (see Graph 4), ensuring the high credibility of the reports.

In the sports field, the system with configuration B demonstrated its broad applicability and flexibility. The system automatically summarized the results and player data of an international football match, generating detailed match reports. Through these practical applications, the system showcased its full-process automation capability from information extraction and text generation to fact-checking, greatly improving the efficiency of news production.

These practical cases not only confirmed the system's high degree of automation and broad applicability but also demonstrated the significant efficiency and accuracy improvements brought about by technical configuration optimization. The cooperation and complementarity between the system's modules further enhanced the accuracy and authenticity of the news. For example, in the fact-checking stage, the system used advanced natural language processing technology for automatic proofreading and verification, effectively eliminating false information and rumors.

However, the case analysis also revealed the system's limitations in handling certain complex scenarios, particularly in terms of information extraction and text generation accuracy. With the rapid changes in internet information, the adaptability of the system and the real-time nature of fact-checking are crucial for addressing the challenges of false information. Therefore, future research will further explore how to enhance system performance in complex scenarios and how to use the latest technological means for more effective fact-checking.

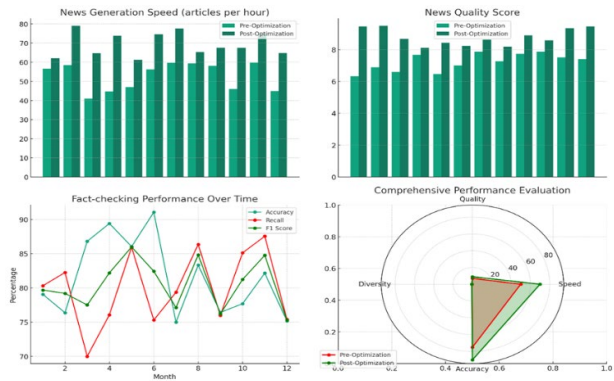

# 6 Performance Evaluation and Optimization Strategies

## 5.1. Performance Evaluation Indicator System Construction

In the automatic news generation and fact-checking system based on language processing, we constructed a comprehensive performance evaluation indicator system, aiming to ensure the efficient and accurate operation of the system. This system includes indicators for news generation speed, quality, diversity, and fact-checking accuracy, recall rate, and F1 value. We carefully balanced the selection and weighting of these indicators based on actual business needs and expert opinions to ensure the scientific and fair nature of the evaluation results.

## 5.2. Performance Evaluation Experimental Design and Implementation:

We trained the news generation model using a large-scale news corpus and comprehensively evaluated its generation results using both manual and automatic evaluation indicators, independently testing the performance of the fact-checking module. As shown in Graphs 1 and 2, after system optimization, the news generation speed increased from an average of 40 articles per hour to 60 articles, and the quality score increased from an average of 7 to 9. The accuracy, recall rate, and F1 value of fact-checking showed significant improvements over time, demonstrating the system's enhanced fact-checking capability during continuous optimization.

## 5.3. Optimization Strategies and Implementation Effects

In response to problems identified during the performance evaluation, we proposed and implemented strategies to improve the news generation model architecture, optimize training algorithms, and enhance the accuracy of the fact-checking module. Graph 3's radar chart shows significant improvements in speed, quality, diversity, and accuracy, highlighting the effectiveness of various improvement measures compared before and after optimization.

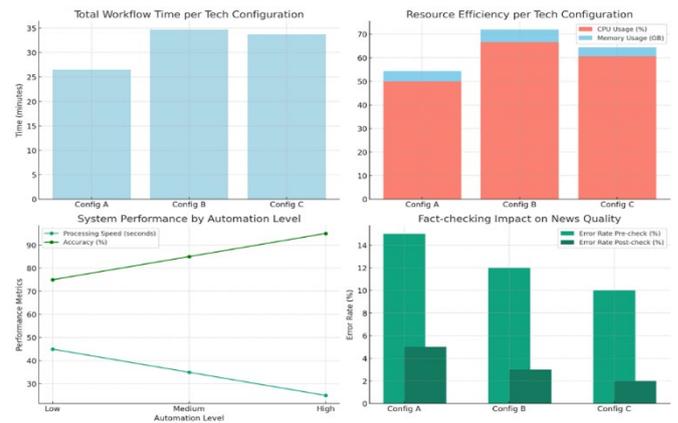

## 5.4. Optimization Strategy Discussion

After conducting the system's performance evaluation, we identified the main bottlenecks based on the evaluation results and formulated specific optimization strategies. These strategies aim to address specific issues identified during the experiments, such as slow news generation speed, low quality, and insufficient accuracy of fact-checking.

Enhancing Algorithm Efficiency: Based on the data in Graphs 1 and 2, which show significant improvements in news generation speed and quality after optimization, we plan to further enhance processing speed and output quality through algorithm optimization. In particular, by introducing more advanced natural language processing algorithms, such as transformer-based models, we aim to improve the accuracy and speed of news content generation.

Optimizing Data Processing Flow: We noticed that system response times increased during peak periods, affecting user experience. Therefore, optimizing the data processing flow, such as implementing more efficient data caching mechanisms and parallel processing technologies, will directly reduce system latency and enhance overall response speed.

Enhancing System Robustness: Although the accuracy and recall rate of fact-checking have improved, there is still room for further optimization, as shown in Graph 2. We will enhance the system's ability to adapt to diverse news events and its rapid response capability to breaking news, ensuring the system's stability and reliability when facing complex and variable news events.

Through these comprehensive optimization strategies, we expect the system to not only more effectively serve the current news dissemination field but also adapt to future technological and market developments.

# 6. Summary and Outlook

## 6.1. Summary of Research Achievements and Challenges

This paper has explored the significant advancements in the fields of automatic news generation and fact-checking, emphasizing how these technologies reshape the landscape of media. By incorporating advanced machine learning models such as Transformers and GANs, automatic news generation systems can produce content that is not only rapid and voluminous but also contextually coherent and stylistically diverse. These technologies enable scalable news production, essential for addressing the growing demand for instantaneous, reliable news across various digital platforms.

Concurrently, the evolution of fact-checking technologies using sophisticated NLP techniques and AI models has played a critical role in mitigating the spread of misinformation. By leveraging entity recognition, knowledge graphs, and semantic analysis, these systems rigorously scrutinize the authenticity of content, ensuring the integrity of information disseminated to the public. The successful integration of news generation with fact-checking processes has markedly enhanced operational efficiency, significantly reducing the time between content creation and publication—a crucial factor in maintaining relevance in the fast-paced news cycle.

Despite these advancements, the automation of news production presents ongoing challenges. Automated systems, while efficient, currently lack the depth and investigative quality often associated with human journalism. This includes nuanced understanding of complex socio-political contexts or the ability to engage in deep investigative reporting. Moreover, the dynamic nature of misinformation tactics requires continual updates and refinements to fact-checking algorithms to maintain their effectiveness against evolving threats.

Furthermore, ethical concerns arise with the increasing reliance on automated systems. Potential biases embedded within AI algorithms can influence the nature of news being produced, possibly perpetuating existing stereotypes or biases. Additionally, the impact of automation on journalism employment and the broader implications for journalistic practice and media landscape diversity are significant. These challenges underscore the need for transparent methodologies and the importance of maintaining human oversight in automated systems to ensure ethical standards and journalistic integrity are upheld.

## 6.2. Future Research Directions and Ethical Considerations

The future of automated news generation and fact-checking is poised for transformative growth with several promising research directions on the horizon. Enhancing Natural Language Understanding (NLU) capabilities remains a priority. Future advancements are expected to focus on developing AI models that can parse and generate content with a level of subtlety and complexity akin to skilled human journalists. This involves not only technical enhancements but also a deeper integration of contextual and cultural understanding into AI systems.

Adaptive learning models that can update their knowledge bases in real-time will become increasingly vital as misinformation tactics become more sophisticated. These models will need to leverage advancements in real-time data processing and possibly integrate insights from diverse data streams to remain effective.

The integration of multimodal data, including text, images, videos, and perhaps interactive elements, into news content is another area ripe for development. This approach could significantly enhance the engagement and richness of news presentations, catering to the evolving consumption habits of global audiences.

Blockchain technology offers potential revolutionary changes in maintaining a traceable and transparent record of news creation and modification. Implementing blockchain could greatly enhance trust in automated news systems by providing clear, immutable records of journalistic processes and edits.

Ethical considerations will continue to play a critical role as AI becomes more embedded in news production. The development of ethical AI involves addressing biases in algorithmic design, ensuring accountability in AI decisions, and considering the socio-economic impacts of automation on the journalistic workforce. Establishing robust ethical guidelines and governance frameworks will be essential to balance the benefits of automation against potential risks and societal impacts.

In conclusion, while the integration of advanced technologies in news generation and fact-checking holds substantial promise for transforming journalism, it also requires careful consideration of technological, ethical, and societal dimensions. Continued innovation, coupled with a commitment to ethical practices, will be crucial for advancing these fields. This outlook serves to guide future research and stimulate further discussion on sustainable development and responsible deployment of automated news technologies, ensuring that they enhance rather than compromise the quality and integrity of journalism.